\documentclass[journal]{IEEEtran}
\usepackage{cite}
\usepackage{amsmath,amssymb,amsfonts}

\DeclareMathOperator*{\argmin}{arg\,min}
\usepackage{algorithmic}
\usepackage{graphicx}
\usepackage{textcomp}
\usepackage{xcolor}
\usepackage[space]{grffile} 
\usepackage{nidanfloat} 
\usepackage{float} 

\usepackage{breqn}
\usepackage{csquotes}
\usepackage{array}
\newcolumntype{P}[1]{>{\centering\arraybackslash}p{#1}}
\newcolumntype{M}{>{\centering\arraybackslash}m{\dimexpr.25\linewidth-2\tabcolsep}}
\newcolumntype{Q}{>{\centering\arraybackslash}m{\dimexpr.5\linewidth-2\tabcolsep}}
\usepackage{placeins}
\usepackage{booktabs}

\renewcommand{\figurename}{Figure }

\newcommand{\subparagraph}{} 
\usepackage{titlesec}

\usepackage{graphicx} 
\graphicspath{{figs/}}

\renewcommand{\tablename}{Table }

\makeatletter
\patchcmd{\@maketitle}
{\addvspace{0.5\baselineskip}\egroup}
{\addvspace{-1\baselineskip}\egroup}
{}
{}
\makeatother

\makeatletter
\newcommand{\linebreakand}{%
\end{@IEEEauthorhalign}
\hfill\mbox{}\par
\mbox{}\hfill\begin{@IEEEauthorhalign}
}

\begin{document}	

\bstctlcite{IEEEexample:BSTcontrol} 
\title{Real-Time, Deep Synthetic Aperture \\ Sonar (SAS) Autofocus}


\author{{Isaac~D.~Gerg and Vishal~Monga}
    \thanks{This work was supported by Office of Naval Research under grants N00014-19-1-2638, N00014-19-1-2513.
    
    I. D. Gerg is with the Applied Research Laboratory and School of EECS at the Pennsylvania State University. V. Monga is with the School of EECS at the Pennsylvania State University (http://signal.ee.psu.edu). The authors thank the Naval Surface Warfare Center - Panama City Division for providing the data used in this experiment.}    
}

\maketitle


\begin{abstract}
Synthetic aperture sonar (SAS) requires precise time-of-flight measurements of the transmitted/received waveform to produce well-focused imagery. It is not uncommon for errors in these measurements to be present resulting in image defocusing. To overcome this, an \emph{autofocus} algorithm is employed as a post-processing step after image reconstruction to improve image focus. A particular class of these algorithms can be framed as a sharpness/contrast metric-based optimization.  To improve convergence, a hand-crafted weighting function to remove ``bad'' areas of the image is sometimes applied to the image-under-test before the optimization procedure. Additionally, dozens of iterations are necessary for convergence which is a large compute burden for low size, weight, and power (SWaP) systems. We propose a deep learning technique to overcome these limitations and implicitly learn the weighting function in a data-driven manner.  Our proposed method, which we call Deep Autofocus, uses features from the single-look-complex (SLC) to estimate the phase correction which is applied in $k$-space. Furthermore, we train our algorithm on batches of training imagery so that during deployment, only a single iteration of our method is sufficient to autofocus.  We show results demonstrating the robustness of our technique by  comparing our results to four commonly used image sharpness metrics. Our results demonstrate Deep Autofocus can produce imagery perceptually better than common iterative techniques but at a lower computational cost. We conclude that Deep Autofocus can provide a more favorable cost-quality trade-off than  alternatives with significant potential of future research.
\end{abstract}

\vspace{-0.5cm}
\section{Introduction}

\begin{figure}[t]
    \centering
    \begin{tabular}{c c}        
        \includegraphics[width=0.35\linewidth]{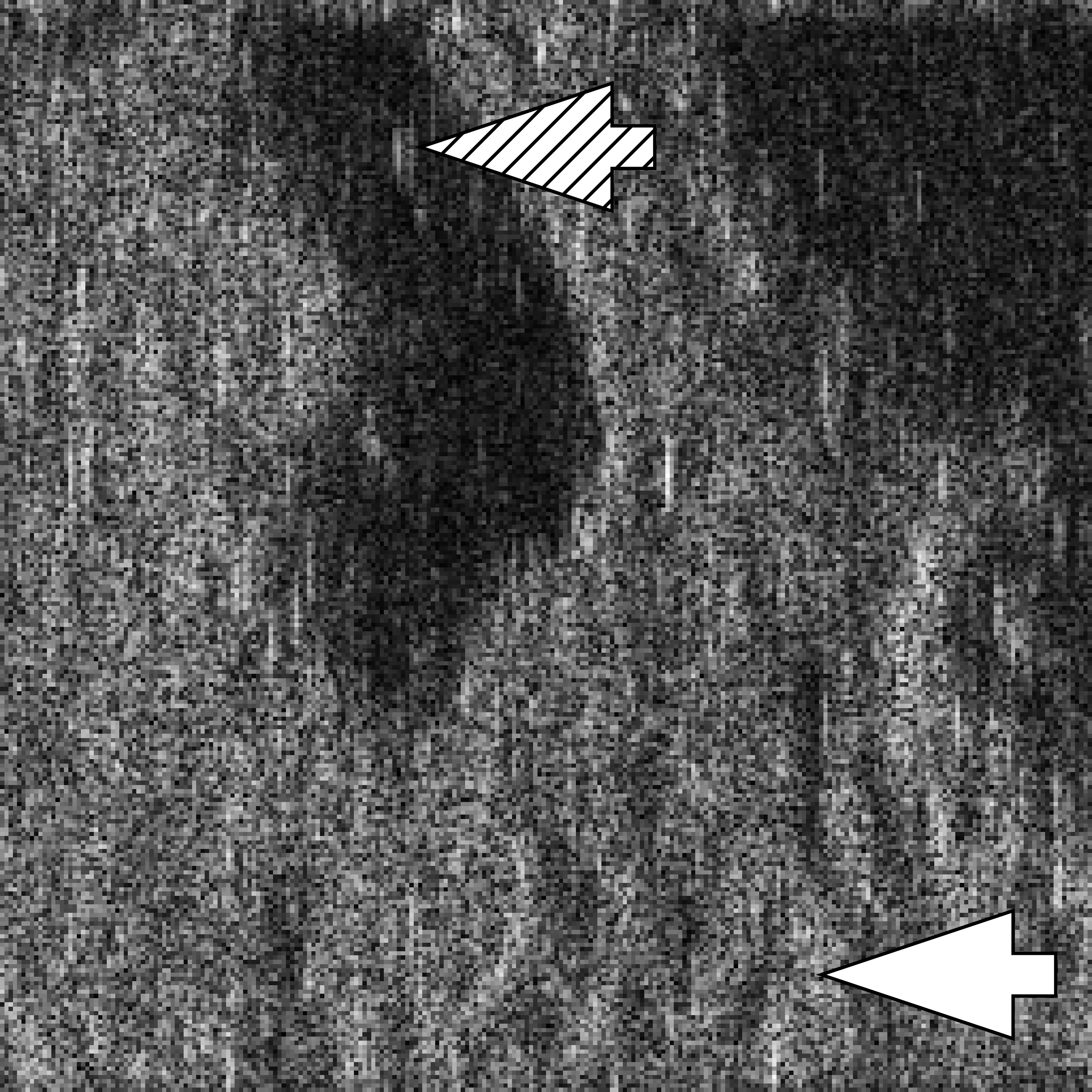} & 
        \includegraphics[width=0.35\linewidth]{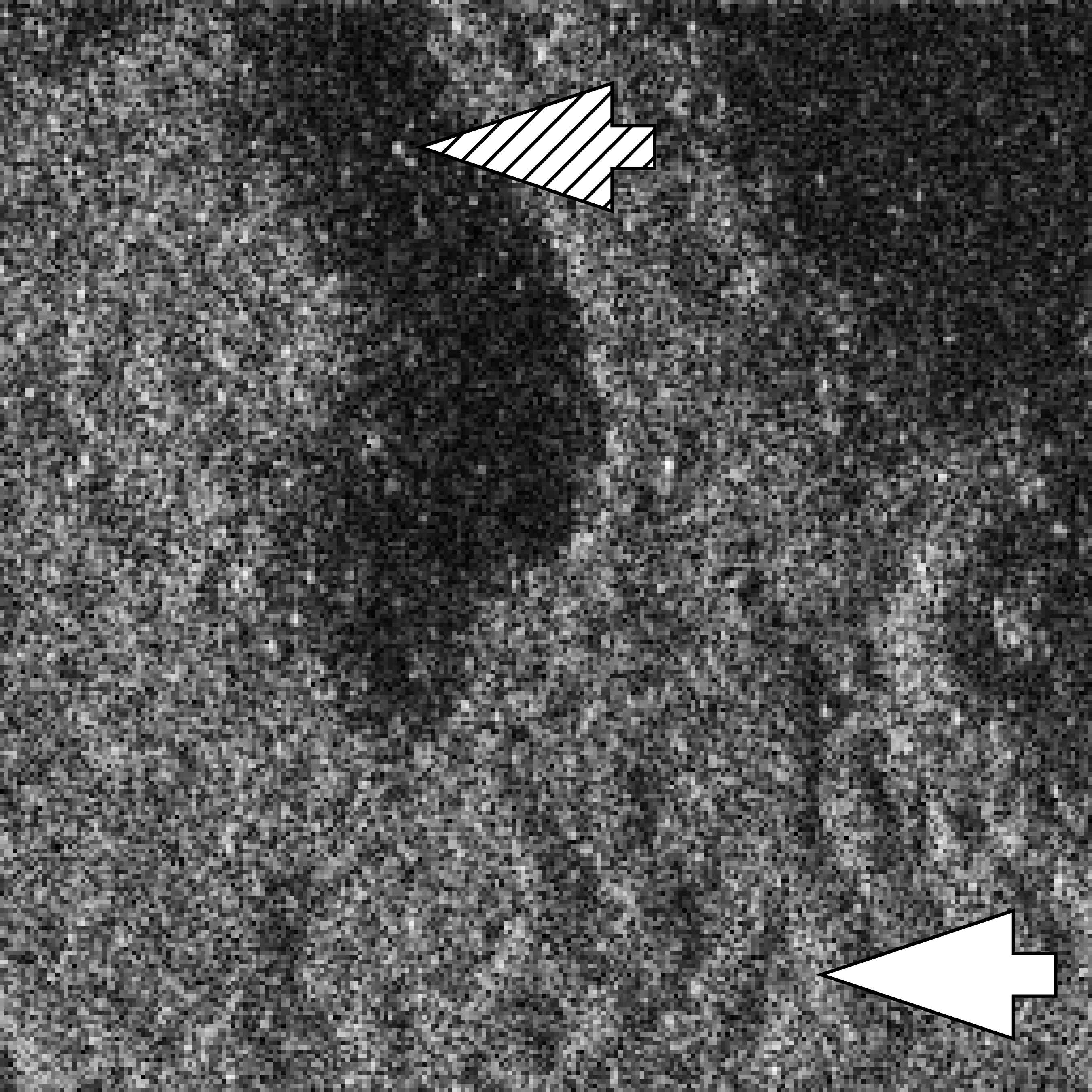} \\
        (a) Original Defocused Image & (b) Autofocus Result
    \end{tabular}
    \caption{(a) Example SAS image with, (b), and without autofocus, (a).  Image (a) is the result of the image reconstruction process. For the image pairs, the arrows show two specific image effects of image defocusing.  The hashed arrow shows how a point scatterer, (b), becomes becomes smeared in the along-track direction (a), which runs vertically in these images.  The performance of some existing autofocus algorithms is dependent on the ability to identify such a point scatter which we see can be non-trivial when blur is large.  The solid arrow shows how seafloor texture can be degraded when image blur, (a), is present.}
    \label{fig:example_images}
\end{figure}

Autofocus for high-frequency (HF) SAS is often employed as a post-processing step after image reconstruction to remove image defocusing.  \figurename \ref{fig:example_images} shows an example SAS defocused/autofocused image pair. There are many error sources which result in image defocusing in SAS \cite{cook2008analysis} such as mis-estimation of sound-speed of vehicle forward velocity.  However, all sources have their root in the incorrect time-of-flight measurement of the transmitted waveform to the seafloor and back to the receive array. A variety of autofocus algorithms exist which are based on an iterative scheme whereby the SLC is modified so that a metric quantifying image sharpness (or contrast) is optimized \cite{callow2003stripmap, wahl1994phase, fienup2000synthetic, fortune2001statistical, fortune2005phase, schulz2006optimal, zeng2013sar, fienup2003aberration}. Recent methods improve upon these by increasing the complexity of the inversion model \cite{li2018coarse, marston2014semiparametric, evers2019generalized, cantalloube2011multiscale}; all have had success. However, many of the algorithms ingest a single image at a time, require several optimization iterations for convergence \cite{fienup2003aberration}, and are vulnerable to converge to local extrema \cite{morrison2003avoiding}.  These factors make for difficult deployment on unmanned underwater vehicles (UUVs) where compute power is at a premium and unreliable autofocus results may confuse the vehicle autonomy engine.  

To make the sharpness-based autofocus optimization procedure converge quickly (i.e. use less compute power) and avoid local extrema (i.e. make more robust), methods often apply a weight map to the image before optimizing \cite{fienup2000synthetic}. The purpose of the map is to remove areas of the image which adversely influence the optimization procedure.  Weighting maps commonly used are designed preserve strong scatters while suppressing image content (e.g. low contrast areas \cite{fienup2000synthetic}) viewed as anomalous with respect to the optimization procedure.

\textbf{Motivation:} Effort has been spent on the design of weighting functions to improve autofocus convergence.  However, humans usually have little issue identifying defocused imagery (and thus detecting poor results out of these iterative autofocus methods).  Our method is inspired by Cook, et al. 2008 \cite{cook2008analysis} whereby the authors show that common phase errors in SAS are easily recognized by trained human operators. 




\textbf{Overview of Contribution:} The autofocus optimization procedure necessitates robust features to converge quickly and to a global extrema.  We design a convolutional neural network (CNN) to automatically learn robust image features from a SAS SLC for the purposes of estimating low-frequency phase error and autofocusing the image. We formulate the optimization so that the compute burden is shifted to the network training phase and autofocus deployment (i.e. inference) is very fast, converging in a single iteration. We do this by training the network with a self-supervised loss not needing ground truth focused/defocused image pairs. During deployment, 1) the CNN extracts features from the dynamic range compressed (DRC) and phase map of the SLC, 2) estimates a phase correction from these features, and 3) applies the phase correction in the $k$-space domain. Consequently, we avoid the need for a hand-crafted weighting function as the method implicitly learns good features from a training database of SAS images.  

\section{Background}
We begin by describing common metric-based iterative autofocus methods \cite{fienup2000synthetic, fienup2003aberration}. We are given a square, well-focused complex-valued SAS image, an SLC, which we denote as $g \in \mathbb{C}^{M \times M}$ where the first dimension is along-track, the second dimension is range, and the sonar transmission arrives on the left side of the image (i.e. $g$ represents a starboard-side collected SLC). We model the defocused image by a spatially uniform phase error throughout the scene represented by
\begin{equation}
    G_e = (e^{i\phi} \otimes \mathbf{1}^T) \odot G 
    \label{eqn:coruption_model}
\end{equation}
where $G$ is the 1-D Fourier transform of $g$ in the along-track dimension (over the image columns) and we denote this as $G = \mathcal{F}\{ g\}$. The phase error over the aperture is $\phi \in \mathbb{R}^{M \times 1}$ and $\mathbf{1}$ is an $M$-element column vector of all ones. $\otimes$ is the Kronecker product (used as a broadcasting operator here) and $\odot$ is the Hadamard product (i.e. pointwise multiplication). The estimated phase error responsible for the image defocusing is $\hat\phi$ and is determined by solving the minimization problem (N.B. maximizing sharpness is minimizing negative sharpness)
\begin{equation}
    \hat{\phi} = \argmin_{\phi} -\mathcal{M}(\mathcal{F}^{-1}\{ (e^{-i\phi} \otimes \mathbf{1}^{T}) \odot G_e \})
    \label{eqn:optimization}
\end{equation}
where $\mathcal{M}$ is one of the sharpness metrics in \tablename \ref{tbl:metrics}. The autofocused image $\hat{g}$ is then given by
\begin{equation}
    \hat{g} = \mathcal{F}^{-1}\{(e^{-i \hat{\phi}} \otimes \mathbf{1}^T) \odot G_e\}
    \label{eqn:apply_autofocus}
\end{equation}

Often, a weighting function, $w \in \mathbb{R}_{+}^{M \times M}$, applied to the argument of $\mathcal{M}$ to remove the influence of unfavorable areas of the image \cite{fienup2000synthetic}.  Accounting for this, the minimization problem becomes
\begin{equation}
    \hat{\phi} = \argmin_{\phi} -\mathcal{M}(w(\vert g_e \vert) \odot \vert \mathcal{F}^{-1}\{ (e^{-i\phi} \otimes \mathbf{1}^T) \odot G_e \vert \})
    \label{eqn:sharpness_optimizaiton}
\end{equation}

Eq \ref{eqn:sharpness_optimizaiton} is solved for each image $g_e$ independently using an iterative method such as gradient descent (GD) or simulated annealing \cite{morrison2003avoiding}. The resultant $\hat{\phi}$ is then applied to $g_e$ using Eq \ref{eqn:apply_autofocus}. Selection of $w$ is determined through a hand-crafted function of the image-under-test; \cite{fienup2000synthetic} gives an example of a common weighting function.

\begin{figure*}[t]
    \centering
    \begin{tabular}{c}        
        \includegraphics[width=0.99\linewidth]{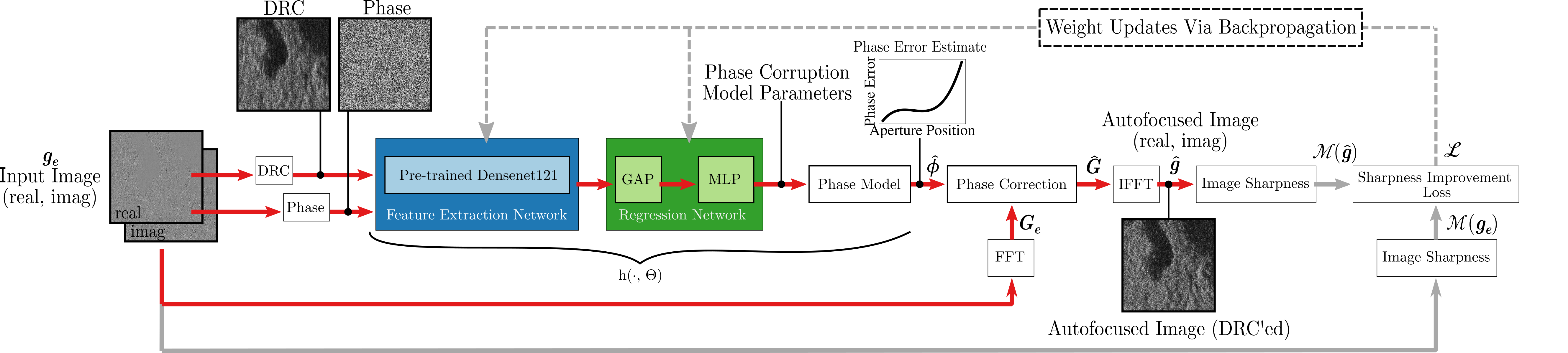}
    \end{tabular}
    \caption{The Deep Autofocus architecture for a mini-batch size of one (recall our mini-batch size is thirty-two during training). The network inputs a single-look complex (SLC) image, $g_e$, which is dynamic range compressed (DRC) and passed into a feature extraction network (blue) along with the SLC's phase.  The features are then fed to a regression network (green) which outputs the phase corruption model parameters, which in this case are the coefficients of ten-degree polynomial modeling the phase error.  The phase error is then applied in the $k$-space domain, computed by the fast Fourier transform (FFT), of the SLC and then inverse Fourier transformed back to the spatial domain.  Finally, the relative improvement in sharpness between the input and output magnitude images is measured and weights are backpropagated to minimize this quantity (recall minimization of this term equates to maximisation of relative image sharpness, see Eq \ref{eqn:rel_function}).  Our formulation is end-to-end differentiable and trained on a graphics processing unit (GPU).  During deployment, only a single forward pass is needed to compute $\hat{g}$ which is represented by the red path. }
    \label{fig:network}
\end{figure*}

\begin{table}[t]
    \centering
    \caption{List of image sharpness metrics used in this work for comparison to Deep Autofocus.  The input single look complex (SLC) is $g \in \mathbb{C}^{M \times M}$ and $\text{stddev}(x)$ is the standard deviation over the elements of $x$.  We make $b = \epsilon$ in OSF as it was shown when $b$ is large the metric is equivalent to SSI.}
    
    \begin{tabular}{c c c}
        \toprule
        \textbf{Ref.} & \textbf{Metric Name} & \textbf{Equation}  \\
        \midrule
        \cite{fortune2001statistical} & Mean Normalized Stddev (MNS) &  $\mathcal{M}_{\text{MNS}} = \frac{\text{stddev}(|g|)}{\text{mean}(|g|)}$  \\
        \cite{zeng2013sar} & Minimum Entropy (ME) & $\mathcal{M}_{\text{ME}} = \sum\limits_{x} \sum\limits_{y} |g|^2 \text{ln}(|g|^2)$ \\
       \cite{schulz2006optimal}  & Optml. Sharpness Function (OSF) & $\mathcal{M}_{\text{OSF}} = \sum\limits_{x} \sum\limits_{y} \text{ln}(|g|^2 + b)$\\            
        \cite{fienup2000synthetic} & Sum of Squared Intensity (SSI) & $\mathcal{M}_{\text{SSI}} = \sum\limits_{x} \sum\limits_{y} |g|^4$  \\
        \bottomrule
    \end{tabular}
    \label{tbl:metrics}
\end{table}

\section{Deep Autofocus}

Deep Autofocus extends the optimization of Eq \ref{eqn:sharpness_optimizaiton} in two ways.  First, we extend the form of $w$ so that it is implicitly learned from a set of training images, specifically from DRC images and phase maps of the SLC. Second, we reformulate the optimization of Eq \ref{eqn:sharpness_optimizaiton} so that during deployment, an iterative method to solve for each image is not needed.  Instead, a fast, single function is applied to all images during deployment.

The goal of Deep Autofocus is to find parameters $\Theta$ for a function $f$ so that 
\begin{equation}
    \hat{g} = f(g_e, \Theta)
\end{equation}
holds for an image $g_e$ selected from a typical population of SAS images.  $\Theta$ is a vector of learned but fixed parameters associated with $f$.  We solve for $\Theta$ by minimization of
\begin{equation}
    \argmin_\Theta \mathcal{L}(g, f(g_e, \Theta))
    \label{eqn:optimization}
\end{equation}
where
\begin{equation}
    f(g_e, \Theta) = \vert \mathcal{F}^{-1} \{ (i \cdot \text{exp}(h(f_{DRC}(g_e), \text{arg}(g_e), \Theta) \otimes \mathbf{1}^T)  \odot G_e \} \vert
\end{equation}
$h$ is a deep convolutional network, $\mathcal{L}$ is the loss function describing the relative sharpness improvement between the input and output image
\begin{equation}
    \mathcal{L}(g_e, \hat{g}) = - \frac{\mathcal{M}(\hat{g}) - \mathcal{M}(g_e)}{\mathcal{M}(g_e)}
    \label{eqn:rel_function}
\end{equation}
where we selected $\mathcal{M}= \mathcal{M}_{MNS}$ (see \tablename \ref{tbl:metrics}), $f_{DRC}$ is the DRC function mapping the SLC to a low dynamic range, human consumable image. $f_{DRC}$ is the rational tone mapping operator of \cite{schlick1995quantization}
\begin{equation}
    f_{DRC}(g) = \frac{q \cdot \vert g \vert}{(q-1) \cdot \vert g \vert + 1} 
\end{equation}
\begin{equation}
    q = \frac{0.2 - 0.2 \cdot \text{median}(\vert g \vert)}{\text{median}(\vert g \vert) - 0.2 \cdot \text{median}(\vert g \vert)}
\end{equation}
We implicitly learn the weighting function $w$ through $h$. Function $h$ takes as input an image and produces features suitable for phase error estimation which is similar to the purpose of $w$. However, $h$ extends $w$ as $w$ is only capable of weighting the image so that ``bad'' areas of the image are suppressed while $h$ is able to do this and selectively enhance or create new features from the image.

The optimization of Eq \ref{eqn:optimization} requires specification of the function family $h$. We use a CNN, DenseNet121 \cite{He_2016_CVPR}, followed by a multi-layer perceptron (MLP) \cite{haykin2007neural}. Densenet121 is composed of 121 layers and serves as a feature extractor generating an output vector in $\mathbb{R}^{8 \times 8 \times 1024}$ which is then dimensionality reduced using global average pooling (GAP) \cite{lin2013network} to $\mathbb{R}^{1024}$.  This vector is fed to the MLP, a sequence of 512-256-128-64-32-8 neurons each followed by leaky rectified linear unit (ReLU) function. The output is a vector in $\mathbb{R}^8$ which are the coefficients representing the low-frequency phase error model, a ten-degree polynomial with degrees zero and one discarded since they have no effect on the sharpness metric. A diagram of our network architecture is shown in \figurename \ref{fig:network}. 

Since Eq \ref{eqn:optimization} (including the Fourier transform and dynamic range compression) is differentiable, stochastic gradient descent (SGD) can be used for optimization to learn $\Theta$ using a small database of training images with data augmentation.  Once training completes, we arrive at the non-iterative function $f$ with fixed, but learned, weights $\Theta$ which estimates the ground truth image $g$ from a potentially defocused image $g_e$. 

To train our network, we use mini-batch size of thirty-two and an SGD learning rate of $10^{-1}$.   We train the model for 10,000 epochs and select for testing the model giving the best validation score. Our training and validation datasets are each composed of 120 images. We employ data augmentation on each training image which is consistent with the method used to generate the test set as described in Section \ref{subsection:dataset_description} . The initial weights, $\Theta$, for the feature extraction network portion of $h$, are from an ImageNet pre-trained Densenet121 model from \cite{tensorflow2015-whitepaper}. For the regression network portion of $h$, the layers are initialized using \cite{glorot2010understanding}.  The model was trained using Tensorflow 2.1 \cite{tensorflow2015-whitepaper} on a graphics processing unit (GPU). 

The point spread function is symmetric for many types of common phase errors (e.g. quadratic phase error) implying the sign of the phase error is not discernible from the DRC image. Thus, phase information is necessary to properly estimate $\phi$.  We verified this by training a network with the phase map input set always to zero and observed suboptimal results.  Additionally, we substituted the DRC and phase map input with a different representation of the SLC, real and imaginary maps, and also observed supobtimal results. Finally, we found optimizing on Eq \ref{eqn:rel_function} instead of directly optimizing $\mathcal{M}(\hat{g})$ gave fastest convergence during training.

\section{Experimental Results and Discussion}
\subsection{Dataset Description}
\label{subsection:dataset_description}
We use a real-world dataset from an HF SAS mounted on a UUV.  The dataset consists of 504 SLC images each 256 $\times$ 256 pixels in size and were constructed using an $\omega$-k beamformer.  The dataset contains seven classes of seafloor: rock, packed sand, mud, small ripple, large ripple, sea grass, and shadow. Of the 504 images, a subset of 264 images are used as test images for algorithm evaluation.  We use these original images as ground truth.  The remaining 240 images are used to train our deep network with half of the images being used for training and half of the images being used for validation. To mimic realistic low-frequency phase error seen in practice \cite{koo2005comparison}, we corrupt each image (see Eq \ref{eqn:coruption_model}) of the test set with phase error from a ten-degree polynomial. This is done by first randomly selecting the order of a polynomial from integer set $\{2,3,...,10\}$.  Next we select the coefficients randomly from $\mathcal{U} [-1,1]$. After that, we normalize the coefficients so that the maximum absolute magnitude of polynomial is 1.0.  Finally, we scale the resultant polynomial by $\mathcal{U}[-18,18]$ radians and apply to the ground-truth SLC. The test images are corrupted once and used for all comparisons. 

\subsection{Evaluation Against Comparison Methods}
We compare our results against four common image sharpness metrics (see \tablename \ref{tbl:metrics}) often used in iterative autofocus. We evaluate the autofocus efficacy and computation run-time performance of each algorithm. 

For autofocus efficacy, we use two common image quality assessment (IQA) metrics: peak-signal-to-noise ratio (PSNR) \cite{wang2004image}  and multi-scale structural similarity (MS-SSIM) \cite{wang2003multiscale}. PSNR is a traditional metric historically used for image comparison. MS-SSIM is contemporary method that correlates well with human assessments of distorted imagery. For each IQA metric, we compare the the original image (the ground truth before corruption with phase error) and the autofocused version we obtain by processing the defocused/corrupted image. To mitigate the effects of speckle, we despeckle the images using \cite{getreuer2012rudin, bush2011bregman} before computing the metric. 

For run-time performance, we measure the time it takes to autofocus all images in the test set. To garner a useful comparison, we allow the sharpness metrics to optimize for ten iterations, likely conservative for deployment in UUV SAS operations. Recall, Deep Autofocus is designed to run using just a single iteration. Each sharpness metric models phase error as a ten-degree polynomial and is minimized using gradient descent (GD).  To garner accurate run-time results, we implemented the sharpness metrics on the same GPU used to run Deep Autofocus.  We did this by implementing the sharpness metrics and the GD procedure on a GPU using Tensorflow. All methods were run on an NVIDIA Titan X. The GD procedure of the sharpness metrics requires a tuning parameter, the learning rate used for GD. To give the best possible results, we used cross-validation to obtain the optimal learning rate for each metric from the set of learning rates $\{ 10^{-6},10^{-5},...,10^{3} \}$.  For each sharpness metric, we selected the learning rate giving the best mean result over the test set.

\begin{figure}[t]  
    \centering
    \begin{tabular}{c}        
        \includegraphics[width=0.95\linewidth]{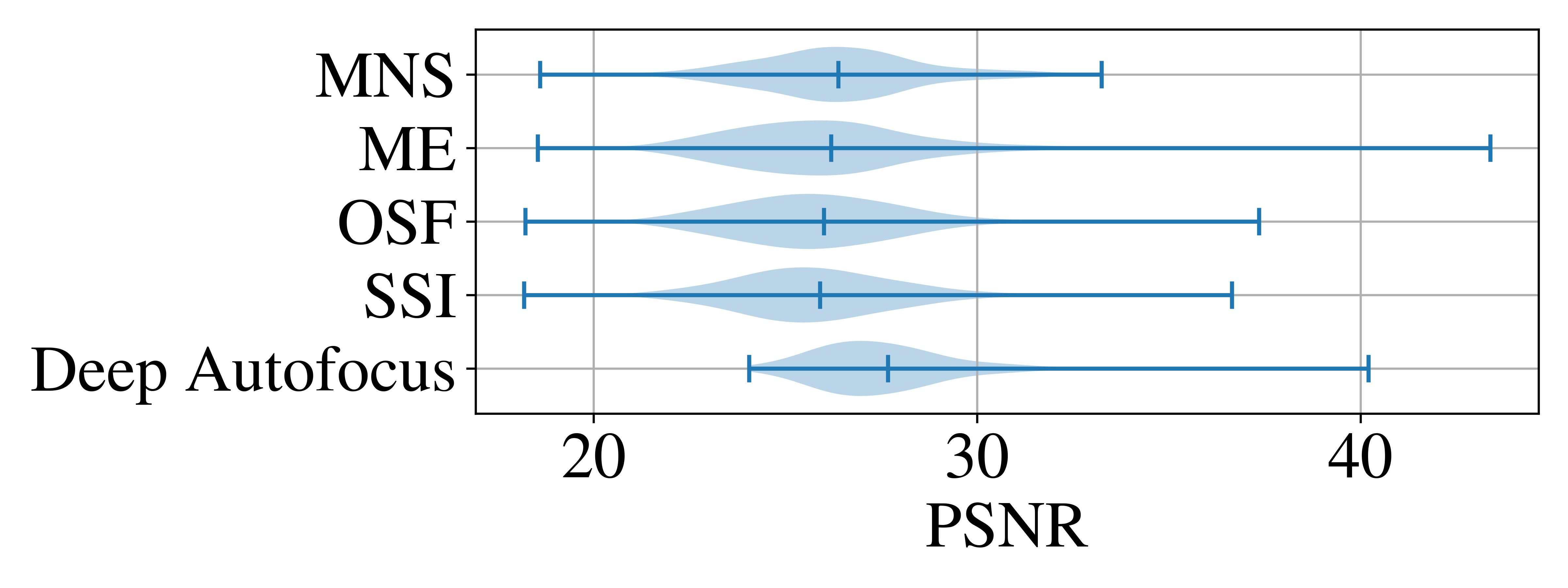} \\
        \includegraphics[width=0.95\linewidth]{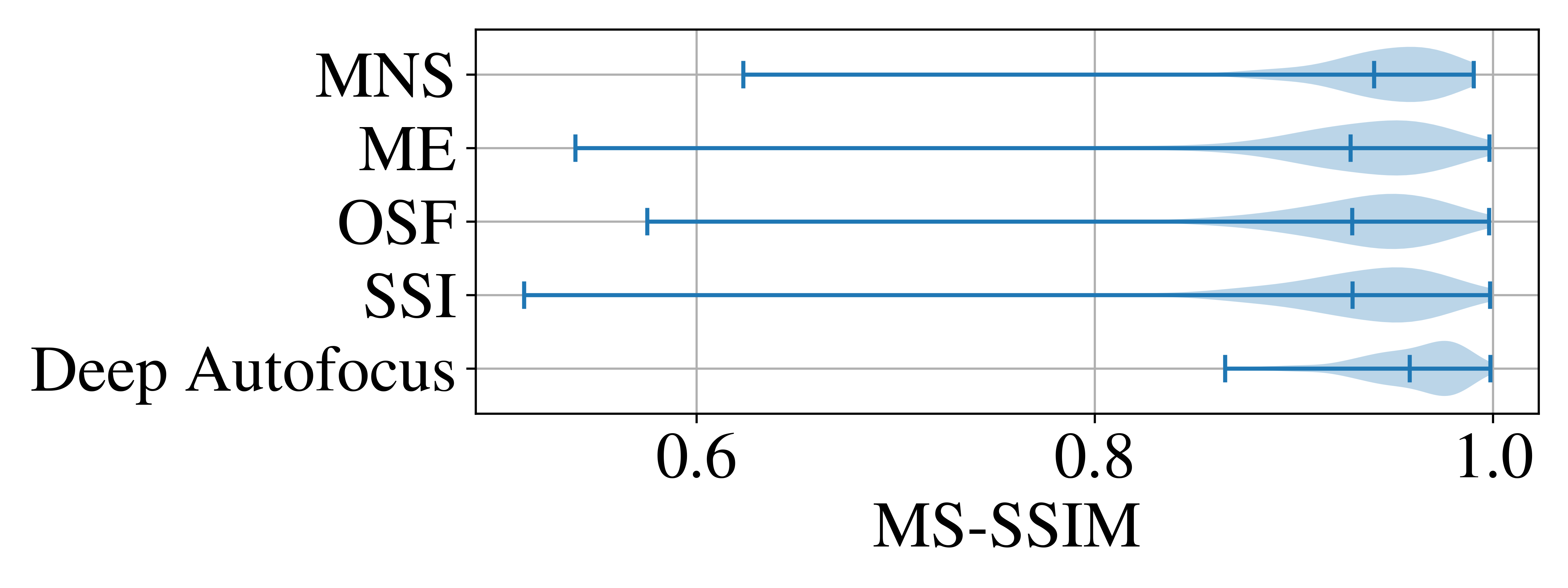} 
    \end{tabular}
    \caption{Image quality comparison of all methods with respect to the 264 ground-truth test images.  We remove speckle noise before computing the metrics using the despeckler of  \cite{getreuer2012rudin, bush2011bregman}. For all metrics, higher values indicated more similarity to the ground truth images. The measures evaluated are peak signal-to-noise ratio (PSNR)\cite{wang2004image} and multiscale structural similarity (MS-SSIM) \cite{wang2003multiscale}. Vertical bars are minimum, mean, and maximum of the distribution.}
    \label{fig:results_psnr_ssim}
\end{figure}

\begin{figure}
    \centering
    \begin{tabular}{c c}
        \includegraphics[width=0.45\linewidth]{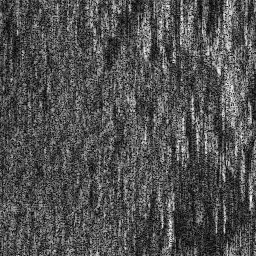} & \includegraphics[width=0.45\linewidth]{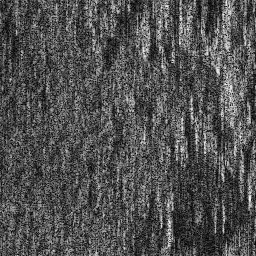} \\
        (a) Image-Under-Test & (b) SSI Autofocus \\
        \includegraphics[width=0.45\linewidth]{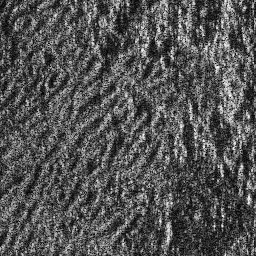} & \includegraphics[width=0.45\linewidth]{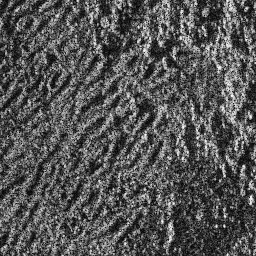}  \\
        (c) Deep Autofocus & (d) Ground Truth\\
    \end{tabular}
    \caption{SSI autofocused image with the worst MS-SSIM and the corresponding Deep Autofocus result.  We can see Deep Autofocus does not catastrophically fail like SSI.  The other three sharpness metrics also failed resulting in similar looking imagery to SSI, (b).}
    \label{fig:worst_msssim}
\end{figure}

We make three observations of our results. First,  Deep Autofocus produces better focused imagery on average by a considerable margin as shown by  \figurename \ref{fig:results_psnr_ssim}.  Second, examining the distributions of \figurename \ref{fig:results_psnr_ssim}, we see Deep Autofocus does not suffer catastrophic failure like the comparison methods; this is visible in the left tail of each violin plot; an example of the behavior is shown in \figurename \ref{fig:worst_msssim}. Finally, the run-time of Deep Autofocus is at least one order of magnitude faster than the comparison methods as shown in \tablename \ref{tbl:runtime}.


\begin{table}[t]
    \centering
    \caption{Run-time performance of each algorithm.  Values were computed by measuring the time it takes to process the test set and then dividing by the number of samples in the test set.  Lower numbers indicate faster run-time.  All metrics optimized using GD exhibit similar run-time.}
    \label{tbl:runtime}        
    \begin{tabular}{c c c}
        \toprule
        \textbf{Algorithm} & \textbf{Iterations} & \textbf{Mean Runtime Per Image [s]} \\
        \midrule
        MNS-GD \cite{fortune2001statistical}  & 10 & $3.4 \times10^{-1}$   \\   
        ME-GD \cite{zeng2013sar}& 10 & $3.4 \times10^{-1}$   \\ 
        OSF-GD \cite{schulz2006optimal} & 10 & $3.4 \times10^{-1}$   \\ 
        SSI-GD \cite{fienup2000synthetic} & 10 & $3.4 \times10^{-1}$   \\ 
        \textbf{Deep Autofocus} & 1 & $\mathbf{1.8 \times10^{-2}}$  \\
        \bottomrule
    \end{tabular}
\end{table}


\bibliographystyle{IEEEtran}
\bibliography{paper_ref}


\onecolumn

\renewcommand{\thesection}{S.\Roman{section}} 
\renewcommand{\thesubsection}{\thesection.\Alph{subsection}}


\end{document}